\documentclass[10pt,twocolumn,letterpaper]{article}

\usepackage{iccv}
\usepackage{times}
\usepackage{epsfig}
\usepackage{graphicx}
\usepackage{amsmath}
\usepackage{amssymb}
\usepackage{booktabs}
\usepackage{multirow}
\usepackage{tabularx}
\usepackage{afterpage}
\usepackage{multicol}

\usepackage[pagebackref=true,breaklinks=true,letterpaper=true,colorlinks,bookmarks=false]{hyperref}

\iccvfinalcopy 

\ificcvfinal\fi
\begin{document}

\title{Multimodal Neurons in Pretrained Text-Only Transformers }

\author{Sarah Schwettmann\textsuperscript{1}\thanks{Indicates equal contribution.}, \hspace{1mm} Neil Chowdhury\textsuperscript{1}\footnotemark[1], \hspace{1mm} Samuel Klein\textsuperscript{2}, \hspace{1mm} David Bau\textsuperscript{3}, \hspace{1mm} Antonio Torralba\textsuperscript{1} \\
\textsuperscript{1}MIT CSAIL, \hspace{1mm} \textsuperscript{2}MIT KFG, \hspace{1mm} \textsuperscript{3}Northeastern University\\
{\tt\small \{schwett, nchow, sjklein, torralba\}@mit.edu, d.bau@northeastern.edu}
}

\maketitle
\ificcvfinal\thispagestyle{empty}\fi

\begin{abstract}
Language models demonstrate remarkable capacity to generalize representations learned in one modality to downstream tasks in other modalities. Can we trace this ability to individual neurons? We study the case where a frozen text transformer is augmented with vision using a self-supervised visual encoder and a single linear projection learned on an image-to-text task. Outputs of the projection layer are not immediately decodable into language describing image content; instead, we find that translation between modalities occurs deeper within the transformer. We introduce a procedure for identifying ``multimodal neurons’' that convert visual representations into corresponding text, and decoding the concepts they inject into the model’s residual stream. In a series of experiments, we show that multimodal neurons operate on specific visual concepts across inputs, and have a systematic causal effect on image captioning. Project page: {\tt\small \href{https://mmns.csail.mit.edu}{mmns.csail.mit.edu}}

\end{abstract}
\vspace{-3mm}
\section{Introduction}


In 1688, William Molyneux posed a philosophical riddle to John Locke that has remained relevant to vision science for centuries: would a blind person, immediately upon gaining sight, visually recognize objects previously known only through another modality, such as touch~\cite{Locke1689-LOCAEC-4, Morgan1977-MORMQV}? A positive answer to the \textit{Molyneux Problem} would suggest the existence a priori of `amodal' representations of objects, common across modalities. In 2011, vision neuroscientists first answered this question in human subjects—\textit{no}, immediate visual recognition is not possible—but crossmodal recognition capabilities are learned rapidly, within days after sight-restoring surgery~\cite{held2011newly}. More recently, language-only artificial neural networks have shown impressive performance on crossmodal tasks when augmented with additional modalities such as vision, using techniques that leave pretrained transformer weights frozen~\cite{tsimpoukelli2021multimodal, eichenberg2021magma, lu2021pretrained, merullo2022linearly, koh2023grounding}.



\begin{figure}[t!]
\begin{center}
\includegraphics[width=\linewidth]{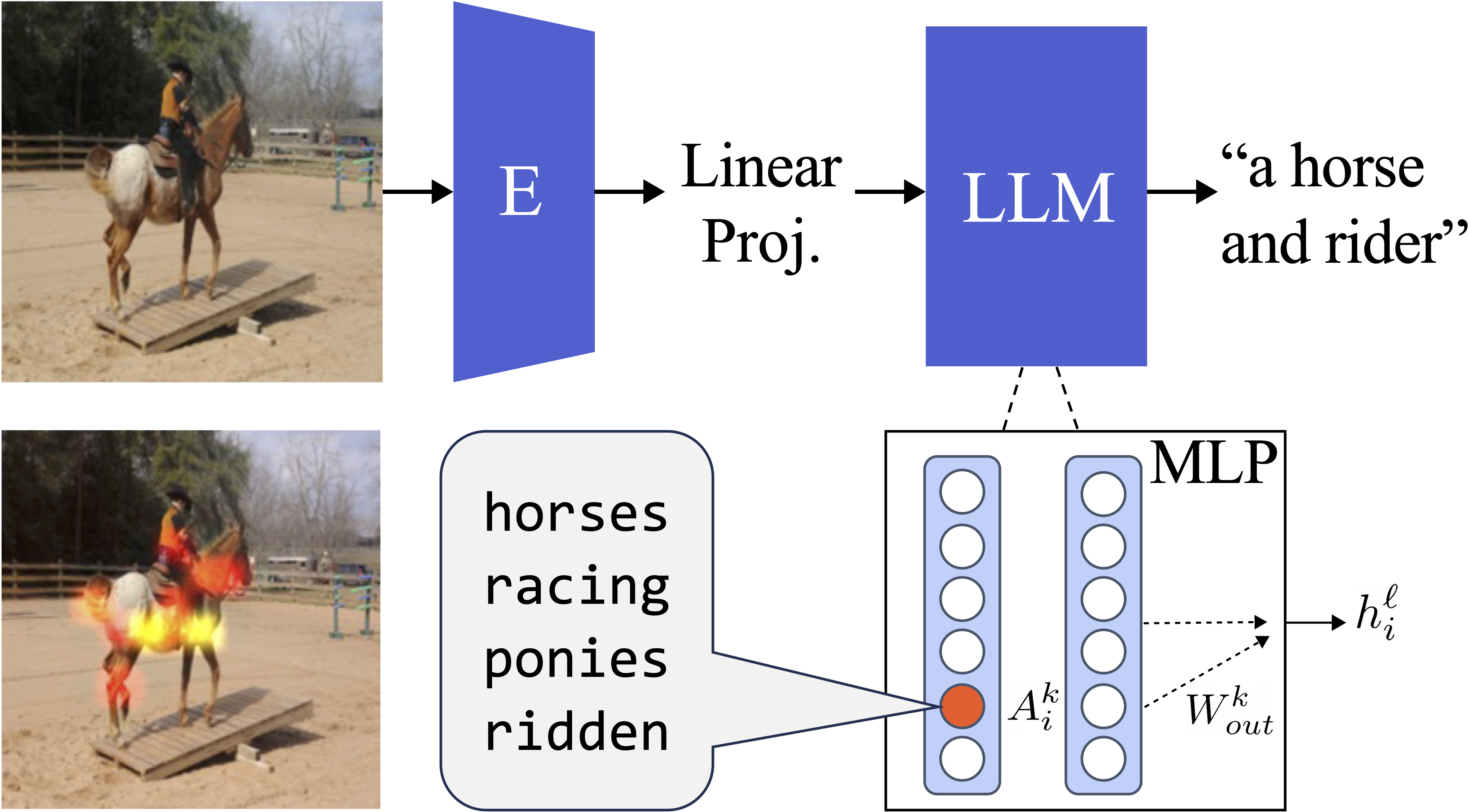}
\vspace{-5.5mm}
\end{center}
   \caption{Multimodal neurons in transformer MLPs activate on specific image features and inject related text into the model's next token prediction. Unit 2019 in GPT-J layer 14 detects horses.}
\label{fig:teaser}
\vspace{-3mm}
\end{figure}


\begin{figure*}[t!]
\begin{center}
\includegraphics[width=\linewidth]{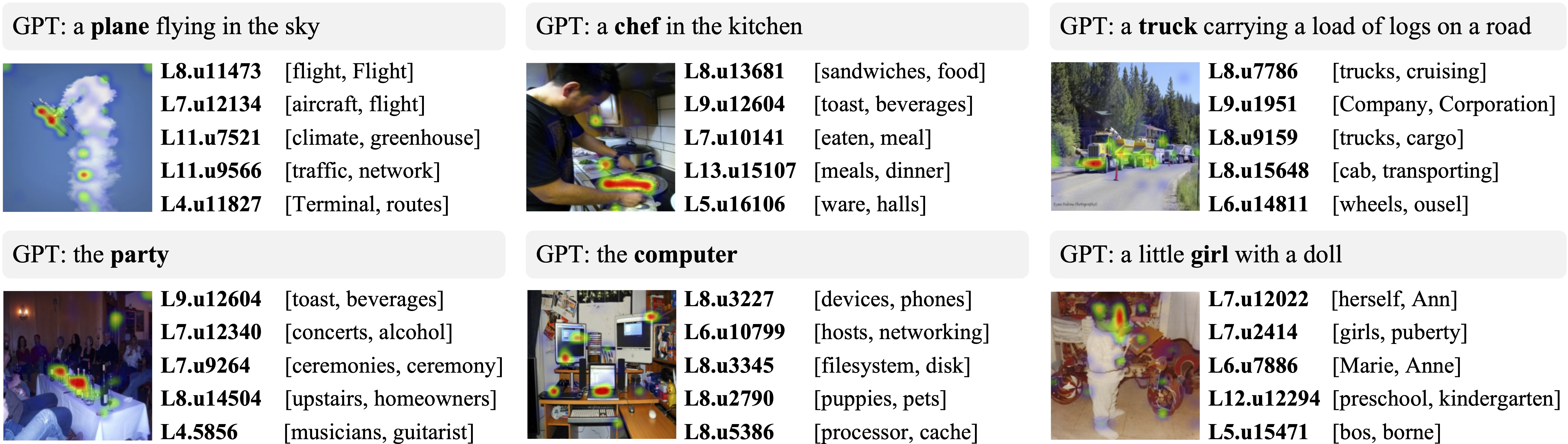}
\end{center}
\vspace{-1mm}
   \caption{Top five multimodal neurons (layer \textbf{L}, unit \textbf{u}), for sample images from 6 COCO supercategories. Superimposed heatmaps ($0.95$ percentile of activations) show mean activations of the top five neurons over the image. Gradient-based attribution scores are computed with respect to the logit shown in bold in the GPT caption of each image. The two highest-probability tokens are shown for each neuron. }
\label{fig:exampleneurons}
\end{figure*}




Vision-language models commonly employ an image-conditioned variant of prefix-tuning \cite{lester2021power, li2021prefix}, where a separate image encoder is aligned to a text decoder with a learned adapter layer. While \textit{Frozen} \cite{tsimpoukelli2021multimodal}, MAGMA \cite{eichenberg2021magma}, and FROMAGe \cite{koh2023grounding} all use image encoders such as CLIP \cite{radford2021learning} trained jointly with language, the recent LiMBeR~\cite{merullo2022linearly} study includes a unique setting: one experiment uses the self-supervised BEIT~\cite{bao2021beit} network, trained with no linguistic supervision, and a linear projection layer between BEIT and GPT-J \cite{gpt-j} supervised by an image-to-text task. This setting is the machine analogue of the Molyneux scenario: the major text components have never seen an image, and the major image components have never seen a piece of text, yet LiMBeR-BEIT demonstrates competitive image captioning performance \cite{merullo2022linearly}. To account for the transfer of semantics between modalities, are visual inputs translated into related text by the projection layer, or does alignment of vision and language representations happen inside the text transformer? In this work, we find: 




\begin{enumerate}
  \item Image prompts cast into the transformer embedding space do not encode interpretable semantics. Translation between modalities occurs inside the transformer. 
  \item Multimodal neurons can be found within the transformer, and they are active in response to particular image semantics.
  \item Multimodal neurons causally affect output: modulating them can remove concepts from image captions.   
\end{enumerate}

\section{Multimodal Neurons}
\label{sec:multimodalneurons}

Investigations of individual units inside deep networks have revealed a range of human-interpretable functions: for example, color-detectors and Gabor filters emerge in low-level convolutional units in image classifiers \cite{erhan2009visualizing}, and later units that activate for object categories have been found across vision architectures and tasks \cite{zeiler2014visualizing, bau2017network, olah2017feature, bau2020units, hernandez2022natural}. \textit{Multimodal neurons} selective for images and text with similar semantics have previously been identified by Goh \etal~\cite{goh2021multimodal} in the CLIP~\cite{radford2021learning} visual encoder, a ResNet-50 model~\cite{he2016deep} trained to align image-text pairs. In this work, we show that multimodal neurons also emerge when vision and language are learned \textit{entirely separately}, and convert visual representations aligned to a frozen language model into text.





\subsection{Detecting multimodal neurons} \label{sec:detect}
We analyze text transformer neurons in the multimodal LiMBeR model \cite{merullo2022linearly}, where a linear layer trained on CC3M~\cite{sharma-etal-2018-conceptual} casts BEIT~\cite{bao2021beit} image embeddings into the input space ($e_{L} = 4096$) of GPT-J 6B~\cite{gpt-j}.
GPT-J transforms input sequence $x = [x_1,\dots, x_P]$ into a probability distribution $y$ over next-token continuations of $x$~\cite{vaswani2017attention}, to create an image caption (where $P=196$ image patches). At layer $\ell$, the hidden state $h_i^\ell$ is given by $h_i^{\ell-1}+\mathbf{a_i}^{\ell}+\mathbf{m_i}^{\ell}$, where $\mathbf{a_i}^{\ell}$ and $\mathbf{m_i}^{\ell}$ are attention and MLP outputs. The output of the final layer $L$ is decoded using $W_d$ for unembedding: $y = \mathrm{softmax}(W_dh^L)$, which we refer to as $~\mathrm{decoder}(h^L)$.

Recent work has found that transformer MLPs encode discrete and recoverable knowledge attributes ~\cite{geva2020transformer, dai2022kneurons, meng2022locating, meng2022massediting}. Each MLP is a two-layer feedforward neural network that, in GPT-J, operates on $h_i^{\ell-1}$ as follows:
\begin{equation} \label{eq:MLP} 
\mathbf{m_i}^{\ell} = W^{\ell}_{out}\mathrm{GELU}(W^{\ell}_{in}h_i^{\ell-1})
\end{equation}
Motivated by past work uncovering interpretable roles of individual MLP neurons in language-only settings \cite{dai2022kneurons}, we investigate their function in a multimodal context. 




\paragraph{Attributing model outputs to neurons with image input.}  
We apply a procedure based on gradients to evaluate the contribution of neuron $u_k$ to an image captioning task. This approach follows several related approaches in neuron attribution, such as Grad-CAM ~\cite{selvaraju2019gradcam} and Integrated Gradients ~\cite{sundararajan2017ig, dai2022kneurons}. We adapt to the recurrent nature of transformer token prediction by attributing generated tokens in the caption to neuron activations, which may be several transformer passes earlier. We assume the model is predicting $c$ as the most probable next token $t$, with logit $y^c$. We define the \textbf{attribution score} $g$ of $u_k$ on token $c$ after a forward pass through image patches $\{1, \dots, p\}$ and pre-activation output $Z$, using the following equation: 

\begin{equation} \label{eq:attribution}   
g_{k, c} = Z^k_p \frac{\partial y^c}{\partial Z^k_p}
\end{equation}

This score is maximized when both the neuron's output and the effect of the neuron are large. It is a rough heuristic, loosely approximating to first-order the neuron's effect on the output logit, compared to a baseline in which the neuron is ablated. Importantly, this gradient can be computed efficiently for all neurons using a single backward pass.

\subsection{Decoding multimodal neurons} 
\label{sec:decodingmultimodalneurons}

What effect do neurons with high $g_{k,c}$ have on model output? We consider $u_k \in U^\ell$, the set of first-layer MLP units ($|U^\ell| = 16384$ in GPT-J). Following Equation~\ref{eq:MLP} and the formulation of transformer MLPs as key-value pairs from~\cite{geva2020transformer}, we note that activation $A_i^k$ of $u_k$ contributes a ``value'' from $W_{out}$ to $h_i$. After the first layer operation:
\begin{equation} \label{eq:decoding} 
\mathbf{m_i} = W_{out}A_i
\end{equation}
As $A_i^k$ grows relative to $A_i^j$ (where $j \neq k$), the direction of $\mathbf{m_i}$ approaches $W_{out}^kA_i^k$, where $W_{out}^k$ is one row of weight matrix $W_{out}$. As this vector gets added to the residual stream, it has the effect of boosting or demoting certain next-word predictions (see Figure \ref{fig:teaser}). To decode the \textit{language contribution} of $u_k$ to model output, we can directly compute $\mathrm{decoder}(W_{out}^k)$, following the simplifying assumption that representations at any layer can be transformed into a distribution over the token vocabulary using the output embeddings \cite{geva2020transformer, geva2022lm, alammar2021ecco,ram2022you}. To evaluate whether $u_k$ translates an image representation into semantically related text, we compare $\mathrm{decoder}(W_{out}^k)$ to image content. 

\begin{table}   
\centering
\begin{tabular}{lccc}
\toprule
& \textbf{BERTScore} &\textbf{CLIPScore} \\
\midrule
random & .3627 & 21.74
\\
multimodal neurons & .3848 & 23.43 
\\
GPT captions & .5251 & 23.62 \\
\bottomrule
\end{tabular}
\vspace{3mm}
\caption{Language descriptions of multimodal neurons correspond with image semantics and  human annotations of images. Scores are reported for a random subset of 1000 COCO validation images. Each BERTScore (F1) is a mean across 5 human image annotations from COCO. For each image, we record the max CLIPScore and BERTScore per neuron, and report means across all images.}
\vspace{-4mm}
\label{tab:neuronevaluation}
\end{table}




\vspace{-3mm}
\paragraph{Do neurons translate image semantics into related text?}  
We evaluate the agreement between visual information in an image and the text multimodal neurons inject into the image caption. 
For each image in the MSCOCO-2017~\cite{lin2014microsoft} validation set, where LiMBeR-BEIT produces captions on par with using CLIP as a visual encoder \cite{merullo2022linearly}, we calculate $g_{k,c}$ for $u_k$ across all layers with respect to the first noun $c$ in the generated caption. For the $100$ $u_k$ with highest $g_{k,c}$ for each image, we compute 
$\mathrm{decoder}(W_{out}^k)$ to produce a list of the $10$ most probable language tokens $u_k$ contributes to the image caption. Restricting analyses to interpretable neurons (where at least 7 of the top 10 tokens are words in the English dictionary containing $\geq$ 3 letters) retains 50\% of neurons with high $g_{k,c}$ (see examples and further implementation details in the Supplement). 






We measure how well decoded tokens (\eg \texttt{horses, racing, ponies, ridden, $\dots$} in Figure \ref{fig:teaser}) correspond with image semantics by computing  CLIPScore~\cite{hessel2021clipscore} relative to the input image and BERTScore~\cite{zhang2019bertscore} relative to COCO image annotations (\eg \textit{a cowboy riding a horse}). Table~\ref{tab:neuronevaluation} shows that tokens decoded from multimodal neurons perform competitively with GPT image captions on CLIPScore, and outperform a baseline on BERTScore where pairings between images and decoded multimodal neurons are randomized (we introduce this baseline as we do not expect BERTScores for comma-separated token lists to be comparable to GPT captions, \eg \textit{a horse and rider}).

Figure~\ref{fig:exampleneurons} shows example COCO images alongside top-scoring multimodal neurons per image, and image regions where the neurons are maximally active. Most top-scoring neurons are found between layers $5$ and $10$ of GPT-J ($L=28$; see Supplement), consistent with the finding from~\cite{meng2022locating} that MLP knowledge contributions occur in earlier layers. 



\begin{figure}[t!]
\begin{center}
\includegraphics[width=\linewidth]{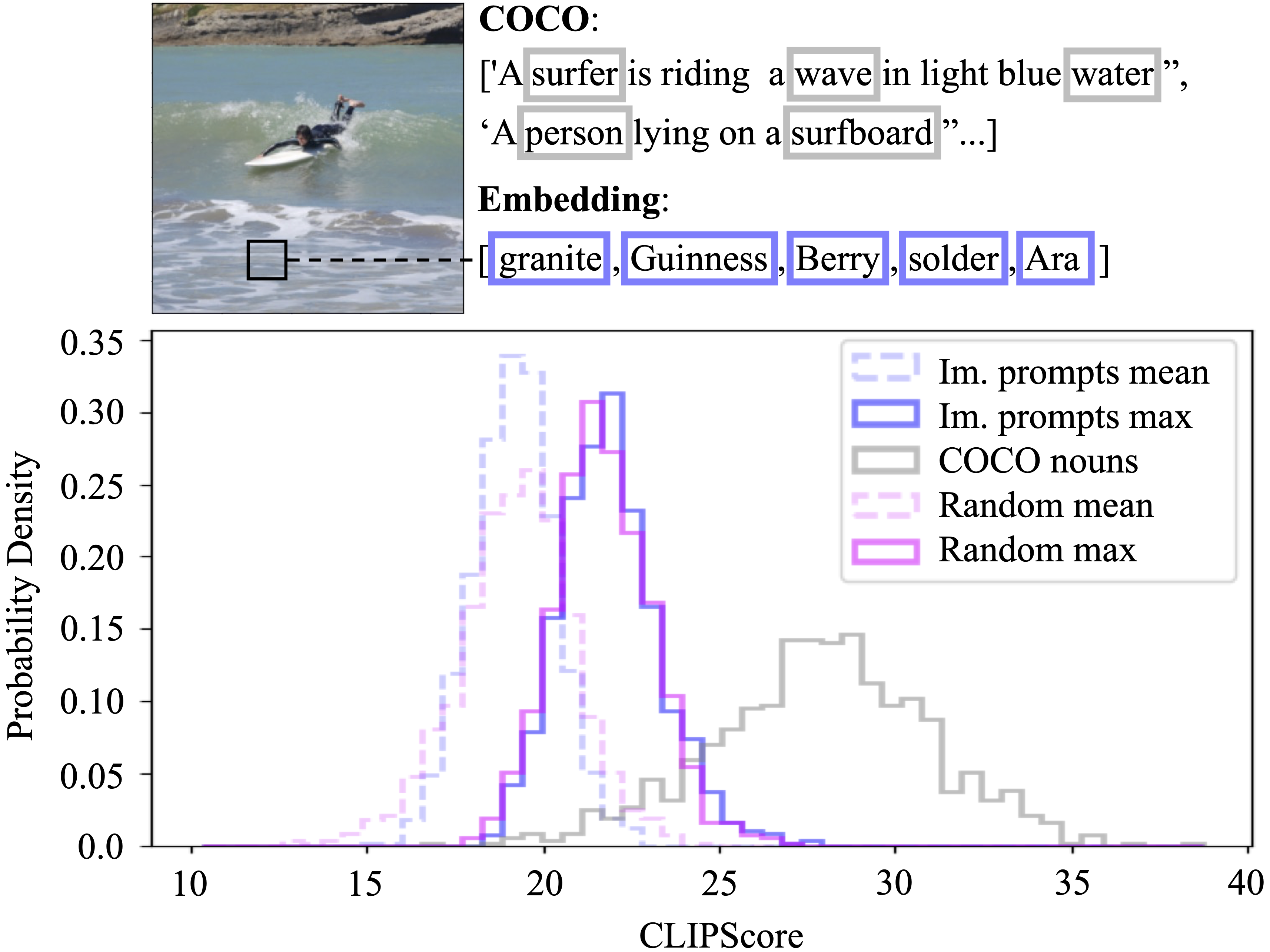}
\end{center}
    \vspace{-4mm}
   \caption{CLIPScores for text-image pairs show no significant difference between decoded image prompts and random embeddings. For image prompts, we report the mean across all image patches as well as the distribution of max CLIPScores per image.}
\label{fig:embeddings}
\end{figure}

\begin{table}[htbp!]
\centering
\begin{tabular}{lcccc}
\toprule
& \textbf{Random} & \textbf{Prompts} &\textbf{GPT} & \textbf{COCO}   \\
\midrule
\textbf{CLIPScore } & 19.22 & 19.17  & 23.62 & \textbf{27.89} \\ 
\textbf{BERTScore} & .3286 & .3291 & \textbf{.5251} & .4470  \\
\bottomrule
\end{tabular}
\vspace{1mm}
\caption{Image prompts are insignificantly different from randomly sampled prompts on CLIPScore and BERTScore. Scores for GPT captions and COCO nouns are shown for comparison.}
\label{tab:embeddings}
\end{table}




\section{Experiments}

\subsection{Does the projection layer translate images into semantically related tokens?}
\label{sec:projection}


We decode image prompts aligned to the GPT-J embedding space into language, and measure their agreement with the input image and its human annotations for 1000 randomly sampled COCO images. As image prompts correspond to vectors in the embedding space and not discrete language tokens, we map them (and 1000 randomly sampled vectors for comparison) onto the five nearest tokens for analysis (see Figure~\ref{fig:embeddings} and Supplement).
A Kolmogorov-Smirnov test~\cite{kolmogorov1933sulla, smirnov1948table} shows no significant difference ($D=.037, p>.5$) between CLIPScore distributions comparing real decoded prompts and random embeddings to images. We compute CLIPScores for five COCO nouns per image (sampled from human annotations) which show significant difference ($D>.9, p<.001$) from image prompts.   


We measure agreement between decoded image prompts and ground-truth image descriptions by computing BERTScores relative to human COCO annotations. Table \ref{tab:embeddings} shows mean scores for real and random embeddings alongside COCO nouns and GPT captions. Real and random prompts are negligibly different, confirming that inputs to GPT-J do not readily encode interpretable semantics.

\subsection{Is visual specificity robust across inputs?}

\begin{figure}[t!]
\begin{center}
\includegraphics[width=\linewidth]{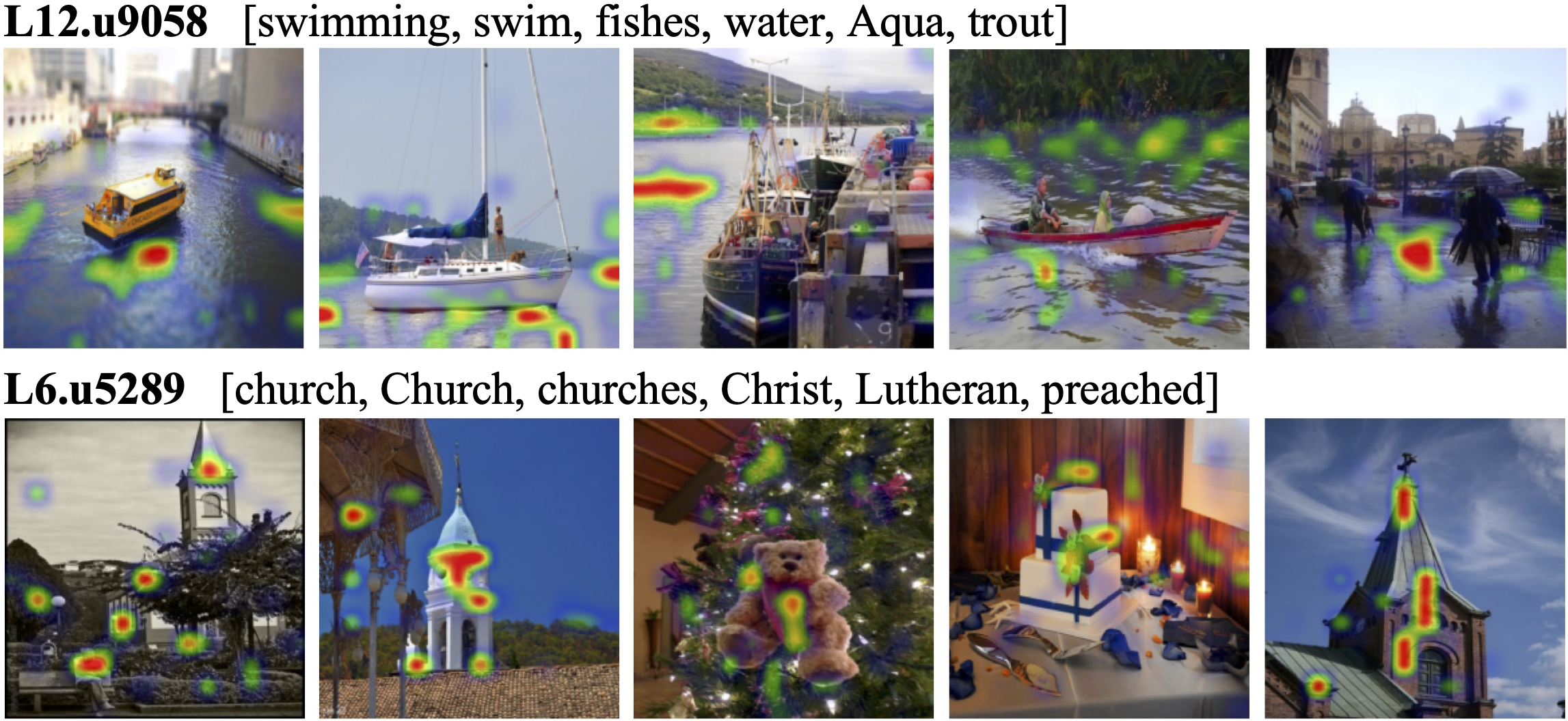}
\end{center}
\vspace{-3mm}
   \caption{Top-activating COCO images for two multimodal neurons. Heatmaps ($0.95$ percentile of activations) illustrate consistent selectivity for image regions translated into related text.}
\label{fig:featureviz}
\end{figure}

A long line of interpretability research has shown that evaluating alignment between individual units and semantic concepts in images is useful for characterizing feature representations in vision models~\cite{netdissect2017, bau2020units, zhou2018interpreting, hernandez2022natural}. Approaches based on visualization and manual inspection (see Figure \ref{fig:featureviz}) can reveal interesting phenomena, but scale poorly.

\begin{figure}[b!]  
\begin{center}
\includegraphics[width=\linewidth]{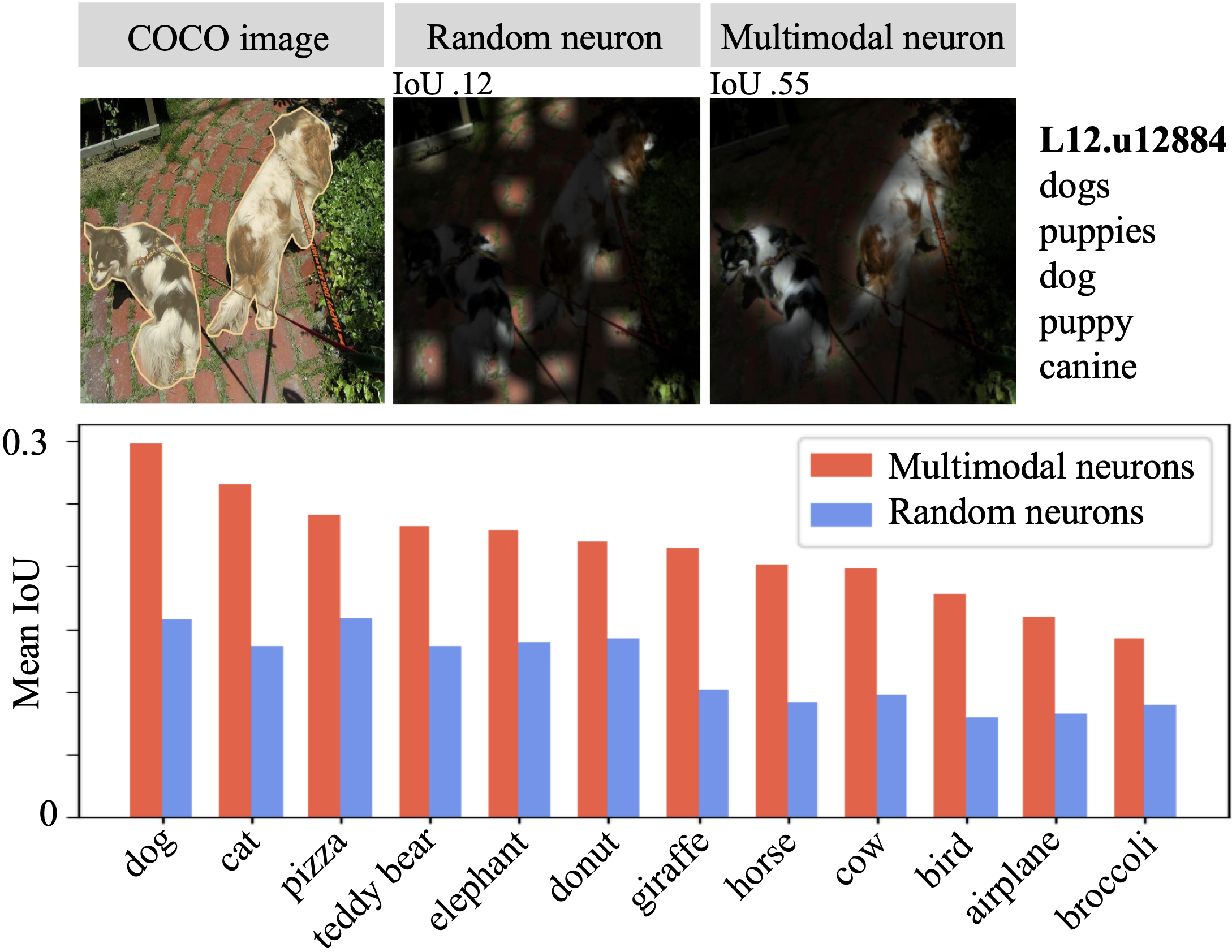}
\end{center}
\vspace{-3mm}
   \caption{Across 12 COCO categories, the receptive fields of multimodal neurons better segment the concept in each image than randomly sampled neurons in the same layers. The Supplement provides additional examples.}
\label{fig:wherelooking}
\end{figure}

We quantify the selectivity of multimodal neurons for specific visual concepts by measuring the agreement of their receptive fields with COCO instance segmentations, following \cite{bau2017network}. 
We simulate the receptive field of $u_k$ by computing $A_i^k$ on each image prompt $x_i \in [x_1,...,x_P]$, reshaping $A_i^k$ into a $14 \times 14$ heatmap, and scaling to $224 \times 224$ using bilinear interpolation. We then threshold activations above the $0.95$ percentile to produce a binary mask over the image, and compare this mask to COCO instance segmentations using Intersection over Union (IoU). To test specificity for individual objects, we select 12 COCO categories with single object annotations, and show that across all categories, the receptive fields of multimodal neurons better segment the object in each image than randomly sampled neurons from the same layers (Figure~\ref{fig:wherelooking}). While this experiment shows that multimodal neurons are reliable detectors of concepts, we also test whether they are selectively active for images containing those concepts, or broadly active across images. Results in the Supplement show preferential activation on particular categories of images.




\begin{figure}[t!]
\begin{center}
\includegraphics[width=\linewidth]{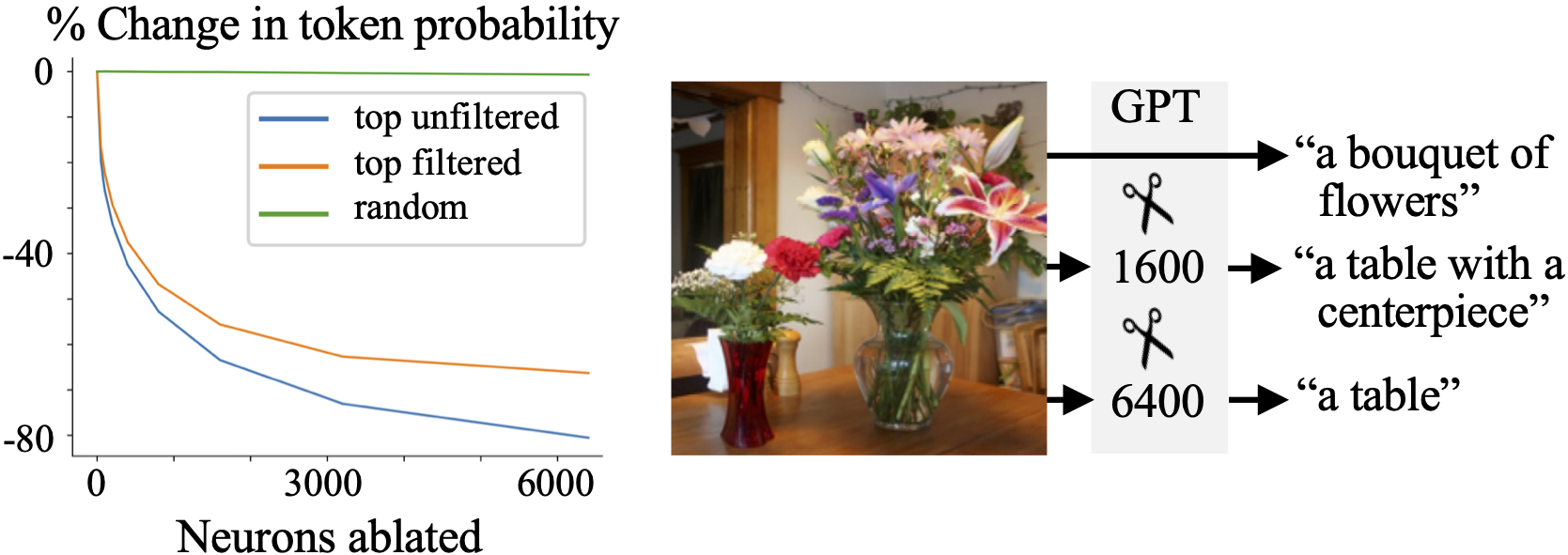}
\end{center}
   \caption{Ablating multimodal neurons degrades image caption content. We plot the effect of ablating multimodal neurons  ordered by $g_{k,c}$ and  randomly sampled units in the same layers (left), and show an example (right) of the effect on a single image caption.}
   \vspace{-4mm}
\label{fig:ablations}
\end{figure}

\subsection{Do multimodal neurons causally affect output?}

To investigate how strongly multimodal neurons causally affect model output, we successively ablate units sorted by $g_{k,c}$ and measure the resulting change in the probability of token $c$. Results for all COCO validation images are shown in Figure \ref{fig:ablations}, for multimodal neurons (filtered and unfiltered for interpretability), and randomly selected units in the same layers. When up to 6400 random units are ablated, we find that the probability of token $c$ is largely unaffected, but ablating the same number of top-scoring units decreases token probability by 80\% on average. Ablating multimodal neurons also leads to significant changes in the semantics of GPT-generated captions. Figure \ref{fig:ablations} shows one example; additional analysis is provided in the Supplement. 

\section{Conclusion}


We find multimodal neurons in text-only transformer MLPs and show that these neurons consistently translate image semantics into language. Interestingly, soft-prompt inputs to the language model do not map onto interpretable tokens in the output vocabulary, suggesting translation between modalities happens \textit{inside} the transformer. The capacity to align representations across modalities could underlie the utility of language models as general-purpose interfaces for 
tasks involving sequential modeling~\cite{lu2021pretrained, hao2022language, srivastava2022beyond, generalpatternmachines2023}, ranging from next-move prediction in games~\cite{li2022emergent, pallagani2022plansformer} to protein design~\cite{unsal2022learning, ferruz2022controllable}. Understanding the roles of individual computational units can serve as a starting point for investigating how transformers generalize across tasks.


%
\section{Limitations}

We study a single multimodal model (LiMBeR-BEIT) of particular interest because the vision and language components were learned separately. The discovery of multimodal neurons in this setting motivates investigation of this phenomenon in other vision-language architectures, and even models aligning other modalities. Do similar neurons emerge when the visual encoder is replaced with a raw pixel stream such as in \cite{lu2021pretrained}, or with a pretrained speech autoencoder?
Furthermore, although we found that the outputs of the LiMBeR-BEIT projection layer are not immediately decodable into interpretable language, our knowledge of the structure of the vector spaces that represent information from different modalities remains incomplete, and we have not investigated how concepts encoded by individual units are assembled from upstream representations. Building a more mechanistic understanding of information processing within transfomers may help explain their surprising ability to generalize to non-textual representations.

\section{Acknowledgements}
We are grateful for the support of the MIT-IBM Watson AI Lab, and ARL grant W911NF-18-2-0218. We thank Jacob Andreas, Achyuta Rajaram, and Tazo Chowdhury for their useful input and insightful discussions.

{\small
\bibliographystyle{ieee_fullname}
\bibliography{egbib}
}

\newpage

\twocolumn[
  \begin{@twocolumnfalse}
    \begin{minipage}{\textwidth}
    \centering
    {\Large\textbf{Supplemental Material for Multimodal Neurons in Pretrained Text-Only Transformers}}
    \end{minipage}
    \vspace{2em}
  \end{@twocolumnfalse}
]
\author{}
\maketitle
\vspace{5mm}
\ificcvfinal\thispagestyle{empty}\fi

\renewcommand{\thefigure}{S.\arabic{figure}}
\setcounter{figure}{0}
\renewcommand{\thetable}{S.\arabic{table}}
\setcounter{table}{0}
\renewcommand{\thesection}{S.\arabic{section}}
\setcounter{section}{0}


\section{Implementation details}
We follow the LiMBeR process for augmenting pretrained GPT-J with vision as described in Merullo \etal (2022). Each image is resized to $(224, 224)$ and encoded into a sequence $[i_1,...,i_k]$ by the image encoder $E$, where $k=196$ and each $i$ corresponds to an image patch of size $(16,16)$. We use self-supervised BEIT as $E$, trained with no linguistic supervision, which produces $[i_1,...,i_k]$ of dimensionality $1024$. To project image representations $i$ into the transformer-defined embedding space of GPT-J, we use linear layer $P$ from Merullo \etal (2022), trained on an image-to-text task (CC3M image captioning). $P$ transforms $[i_1,...,i_k]$ into soft prompts $[x_1,...,x_k]$ of dimensionality $4096$, which we refer to as the image prompt. Following convention from SimVLM, MAGMA and LiMBeR, we append the text prefix ``A picture of'' after every every image prompt. Thus for each image, GPT-J receives as input a $(199,4096)$ prompt and outputs a probability distribution $y$ over next-token continuations of that prompt.

To calculate neuron attribution scores, we generate a caption for each image by sampling from $y$ using temperature $T=0$, which selects the token with the highest probability at each step. The attribution score $g_{k,c}$ of neuron $k$ is then calculated with respect to token $c$, where $c$ is the first noun in the generated caption (which directly follows the image prompt and is less influenced by earlier token predictions). In the rare case where this noun is comprised of multiple tokens, we let $c$ be the first of these tokens. This attribution score lets us rank multimodal neurons by how much they contribute to the crossmodal image captioning task.

\section{Example multimodal neurons}
\label{sec:ExampleImages}

Table S.1 shows additional examples of multimodal neurons detected and decoded for randomly sampled images from the COCO 2017 validation set. The table shows the top 20 neurons across all MLP layers for each image. In analyses where we filter for interpretable neurons that correspond to objects or object features in images, we remove neurons that decode primarily to word fragments or punctuation. Interpretable units (units where at least 7 of the top 10 tokens are words in the SCOWL English dictionary, for en-US or en-GB, with $\geq 3$ letters) are highlighted in bold.

\section{Evaluating agreement with image captions}
We use BERTScore (F1) as a metric for evaluating how well a list of tokens corresponds to the semantic content of an image caption. Section 2.2 uses this metric to evaluate multimodal neurons relative to ground-truth human annotations from COCO, and Section 3.1 uses the metric to determine whether projection layer $P$ translates $[i_1,...,i_k]$ into $[x_1,...,x_k]$ that already map visual features onto related language before reaching transformer MLPs. Given that $[x_1,...,x_k]$ do not correspond to discrete tokens, we map each $x$ onto the $5$ token vectors with highest cosine similarity in the transformer embedding space for analysis. 

Table S.2 shows example decoded soft prompts for a randomly sampled COCO image. For comparison, we sample random vectors of size $4096$ and use the same procedure to map them onto their nearest neighbors in the GPT-J embedding space. BERTScores for the random soft prompts are shown alongside scores for the image soft prompts. The means of these BERTScores, as well as the maximum values, are indistinguishable for real and random soft prompts (see Table S.2 for a single image and Figure 3 in the main paper for the distribution across COCO images). Thus we conclude that $P$ produces image prompts that fit within the GPT-J embedding space, but do not already map image features onto related language: this occurs deeper inside the transformer.

\section{Selectivity of multimodal neurons}
Figure S.1 shows additional examples of activation masks of individual multimodal neurons over COCO validation images, and IoU scores comparing each activation mask with COCO object annotations.

We conduct an additional experiment to test whether multimodal neurons are selectively active for images containing particular concepts. If unit $k$ is selective for the images it describes (and not, for instance, for many images), then we expect greater $A^k_{x_i}$ on images where it relevant to the caption than on images where it is irrelevant. It is conceivable that our method merely extracts a set of high-activating neurons, not a set of neurons that are selectively active on the inputs we claim they are relevant to captioning.

We select 10 diverse ImageNet classes (see Figure S.2) and compute the top 100 scoring units per image on each of 200 randomly sampled images per class in the ImageNet training set, filtered for interpretable units. Then for each class, we select the 20 units that appear in the most images for that class. We measure the mean activation of these units across all patches in the ImageNet validation images for each of the 10 classes. Figure S.2(a) shows the comparison of activations across each of the categories. We find that neurons activate more frequently on images in their own category than for others. This implies that our pipeline does not extract a set of general visually attentive units, but rather units that are specifically tied to image semantics.

\begin{table*}[htbp]
\centering
\small
\label{table:example}
\begin{tabularx}{\linewidth}{c l l p{9.6cm} r}
\toprule
\textbf{Images} & \textbf{Layer.unit} & \textbf{Patch} & \textbf{Decoding (top 5 tokens)} & \textbf{Attr. score} \\
\midrule
\multirow{6}{*}{\includegraphics[width=0.115\linewidth]{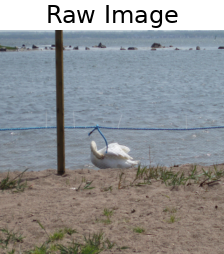}} & \textbf{L7.u15772} & \textbf{119} & \textbf{` animals', ` embryos', ` kittens', ` mammals', ` eggs'} & \textbf{0.0214} \\
 & \textbf{L5.u4923} & \textbf{119} & \textbf{` birds', ` cages', ` species', ` breeding', ` insects'} & \textbf{0.0145} \\
 & \textbf{L7.u12134} & \textbf{119} & \textbf{` aircraft', ` flight', ` airplanes', ` Flight', ` Aircraft'} & \textbf{0.0113} \\
 & \textbf{L5.u4888} & \textbf{119} & \textbf{` Boat', ` sails', `voy', ` boats', ` ships'} & \textbf{0.0085} \\
 & \textbf{L7.u5875} & \textbf{119} & \textbf{` larvae', ` insects', ` mosquitoes', ` flies', ` species'} & \textbf{0.0083} \\
 & \textbf{L8.u2012} & \textbf{105} & \textbf{` whales', ` turtles', ` whale', ` birds', ` fishes'} & \textbf{0.0081} \\
 & \textbf{L7.u3030} & \textbf{119} & \textbf{` Island', ` island', ` Islands', ` islands', ` shore'} & \textbf{0.0078} \\
\multirow{6}{*}{\includegraphics[width=0.115\linewidth]{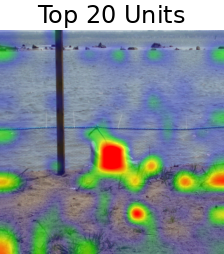}} & L7.u14308 & 119 & `uses', ` dec', `bill', `oid', `FS' & 0.0078 \\
 & \textbf{L9.u12771} & \textbf{119} & \textbf{` satellites', ` Flight', ` orbiting', ` spacecraft', ` ship'} & \textbf{0.0075} \\
 & \textbf{L4.u12317} & \textbf{119} & \textbf{` embryos', ` chicken', ` meat', ` fruits', ` cows'} & \textbf{0.0071} \\
 & \textbf{L8.u2012} & \textbf{119} & \textbf{` whales', ` turtles', ` whale', ` birds', ` fishes'} & \textbf{0.0062} \\
 & \textbf{L5.u4530} & \textbf{119} & \textbf{` herds', ` livestock', ` cattle', ` herd', ` manure'} & \textbf{0.0056} \\
 & \textbf{L5.u4923} & \textbf{105} & \textbf{` birds', ` cages', ` species', ` breeding', ` insects'} & \textbf{0.0055} \\
 & \textbf{L6.u8956} & \textbf{119} & \textbf{` virus', ` strains', ` infect', ` viruses', ` parasites'} & \textbf{0.0052} \\
\multirow{6}{*}{\includegraphics[width=0.115\linewidth]{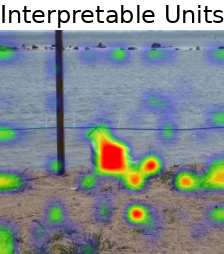}} & L7.u2159 & 105 & ` species', `species', ` bacteria', ` genus', ` Species' & 0.0051 \\
 & L10.u4819 & 119 & `çĶ°', `¬¼', `"""', ` Marketable', `å§' & 0.0051 \\
 & \textbf{L5.u4923} & \textbf{118} & \textbf{` birds', ` cages', ` species', ` breeding', ` insects'} & \textbf{0.0050} \\
 & L10.u927 & 3 & `onds', `rog', `lys', `arrow', `ond' & 0.0050 \\
 & \textbf{L11.u7635} & \textbf{119} & \textbf{` birds', `birds', ` butterflies', ` kittens', ` bird'} & \textbf{0.0049} \\
 & \textbf{L9.u15445} & \textbf{119} & \textbf{` radar', ` standby', ` operational', ` flight', ` readiness'} & \textbf{0.0048} \\
\midrule
\multirow{6}{*}{\includegraphics[width=0.115\linewidth]{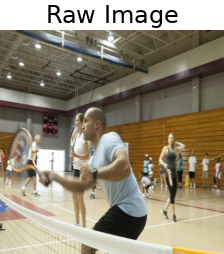}} & \textbf{L5.u15728} & \textbf{119} & \textbf{` playoff', ` players', ` teammate', ` player', `Players'} & \textbf{0.0039} \\
 & L12.u11268 & 113 & `elson', `ISA', `Me', `PRES', `SO' & 0.0039 \\
 & \textbf{L5.u9667} & \textbf{119} & \textbf{` workouts', ` workout', ` Training', ` trainer', ` exercises'} & \textbf{0.0034} \\
 & L9.u15864 & 182 & `lihood', `/**', `Advertisements', `.''.', `"""' & 0.0034 \\
 & \textbf{L9.u9766} & \textbf{119} & \textbf{` soccer', ` football', ` player', ` baseball', `player'} & \textbf{0.0033} \\
 & L10.u4819 & 182 & `çĶ°', `¬¼', `"""', ` Marketable', `å§' & 0.0033 \\
 & L18.u15557 & 150 & `imer', `ohan', `ellow', `ims', `gue' & 0.0032 \\
\multirow{6}{*}{\includegraphics[width=0.115\linewidth]{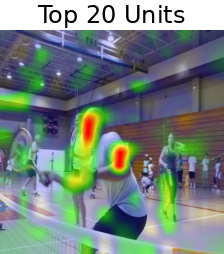}} & L12.u6426 & 160 & `â¢', ` Â®', ` syndrome', ` Productions', ` Ltd' & 0.0032 \\
 & \textbf{L8.u15435} & \textbf{119} & \textbf{` tennis', ` tournaments', ` tournament', ` golf', ` racing'} & \textbf{0.0032} \\
 & \textbf{L11.u4236} & \textbf{75} & \textbf{` starring', ` played', ` playable', ` Written', ` its'} & \textbf{0.0031} \\
 & \textbf{L8.u6207} & \textbf{119} & \textbf{` player', ` players', ` Player', `Ä', ` talent'} & \textbf{0.0031} \\
 & \textbf{L6.u5975} & \textbf{119} & \textbf{` football', ` soccer', ` basketball', ` Soccer', ` Football'} & \textbf{0.0030} \\
 & L2.u10316 & 75 & `ï', `/**', `Q', `The', `//' & 0.0028 \\
 & L12.u8390 & 89 & `etheless', `viously', `theless', `bsite', `terday' & 0.0028 \\
\multirow{6}{*}{\includegraphics[width=0.115\linewidth]{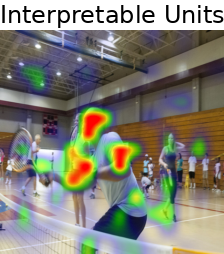}} & \textbf{L5.u7958} & \textbf{89} & \textbf{` rugby', ` football', ` player', ` soccer', ` footballer'} & \textbf{0.0028} \\
 & L20.u9909 & 89 & ` Associates', ` Alt', ` para', ` Lt', ` similarly' & 0.0026 \\
 & \textbf{L5.u8219} & \textbf{75} & \textbf{` portion', ` regime', ` sector', ` situation', ` component'} & \textbf{0.0026} \\
 & \textbf{L11.u7264} & \textbf{75} & \textbf{` portion', ` finale', ` environment', `iest', ` mantle'} & \textbf{0.0026} \\
 & L20.u452 & 103 & ` CLE', ` plain', ` clearly', ` Nil', ` Sullivan' & 0.0026 \\
 & L7.u16050 & 89 & `pc', `IER', ` containing', ` formatted', ` supplemented' & 0.0026 \\
\midrule
\multirow{6}{*}{\includegraphics[width=0.115\linewidth]{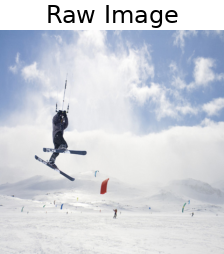}} & L10.u927 & 73 & `onds', `rog', `lys', `arrow', `ond' & 0.0087 \\
 & \textbf{L5.u9667} & \textbf{101} & \textbf{` workouts', ` workout', ` Training', ` trainer', ` exercises'} & \textbf{0.0081} \\
 & L9.u3561 & 73 & ` mix', ` CRC', ` critically', ` gulf', ` mechanically' & 0.0076 \\
 & \textbf{L9.u5970} & \textbf{73} & \textbf{` construct', ` performance', ` global', ` competing', ` transact'} & \textbf{0.0054} \\
 & L10.u562 & 73 & ` prev', ` struct', ` stable', ` marg', ` imp' & 0.0054 \\
 & \textbf{L6.u14388} & \textbf{87} & \textbf{` march', ` treadmill', ` Championships', ` racing', ` marathon'} & \textbf{0.0052} \\
 & \textbf{L14.u10320} & \textbf{73} & \textbf{` print', ` handle', ` thing', `catch', `error'} & \textbf{0.0051} \\
\multirow{6}{*}{\includegraphics[width=0.115\linewidth]{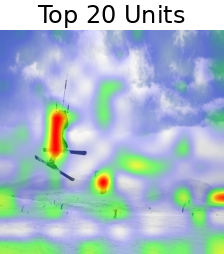}} & L9.u3053 & 73 & `essel', `ked', ` ELE', `ument', `ue' & 0.0047 \\
 & L5.u4932 & 73 & `eman', `rack', `ago', `anne', `ison' & 0.0046 \\
 & L9.u7777 & 101 & `dr', `thur', `tern', `mas', `mass' & 0.0042 \\
 & L6.u16106 & 73 & `umble', `archives', `room', ` decentral', `Root' & 0.0040 \\
 & \textbf{L5.u14519} & \textbf{73} & \textbf{` abstract', ` global', `map', `exec', `kernel'} & \textbf{0.0039} \\
 & L11.u10405 & 73 & `amed', `elect', `1', `vol', `vis' & 0.0038 \\
 & L9.u325 & 87 & ` training', ` tournaments', `ango', ` ballet', ` gymn' & 0.0038 \\
\multirow{6}{*}{\includegraphics[width=0.115\linewidth]{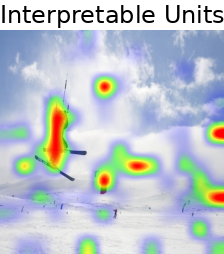}} & \textbf{L6.u14388} & \textbf{101} & \textbf{` march', ` treadmill', ` Championships', ` racing', ` marathon'} & \textbf{0.0038} \\
 & L7.u3844 & 101 & `DERR', `Charges', `wana', `¬¼', `verages' & 0.0036 \\
 & L9.u15864 & 101 & `lihood', `/**', `Advertisements', `.''.', `"""' & 0.0036 \\
 & \textbf{L7.u3330} & \textbf{101} & \textbf{` Officers', ` officers', ` patrolling', ` patrols', ` troops'} & \textbf{0.0036} \\
 & \textbf{L8.u8807} & \textbf{73} & \textbf{` program', ` updates', ` programs', ` document', ` format'} & \textbf{0.0034} \\
 & \textbf{L6.u12536} & \textbf{87} & \textbf{` ankles', ` joints', ` biome', ` injuries', ` injury'} & \textbf{0.0034} \\
\midrule
\end{tabularx}
\end{table*}

\begin{table*}[htbp]
\centering
\small
\label{table:example2}
\begin{tabularx}{\linewidth}{c l l p{9.6cm} r}
\toprule
\textbf{Images} & \textbf{Layer.unit} & \textbf{Patch} & \textbf{Decoding (top 5 tokens)} & \textbf{Attr. score} \\
\midrule
\multirow{6}{*}{\includegraphics[width=0.115\linewidth]{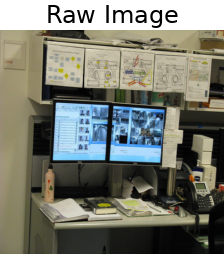}} & \textbf{L8.u14504} & \textbf{13} & \textbf{` upstairs', ` homeowners', ` apartments', ` houses', ` apartment'} & \textbf{0.0071} \\
 & \textbf{L13.u15107} & \textbf{93} & \textbf{` meals', ` meal', ` dinner', ` dishes', ` cuisine'} & \textbf{0.0068} \\
 & \textbf{L8.u14504} & \textbf{93} & \textbf{` upstairs', ` homeowners', ` apartments', ` houses', ` apartment'} & \textbf{0.0052} \\
 & \textbf{L8.u14504} & \textbf{150} & \textbf{` upstairs', ` homeowners', ` apartments', ` houses', ` apartment'} & \textbf{0.0048} \\
 & \textbf{L9.u4691} & \textbf{13} & \textbf{` houses', ` buildings', ` dwellings', ` apartments', ` homes'} & \textbf{0.0043} \\
 & \textbf{L8.u13681} & \textbf{93} & \textbf{` sandwiches', ` foods', ` salad', ` sauce', ` pizza'} & \textbf{0.0041} \\
 & \textbf{L12.u4638} & \textbf{93} & \textbf{` wash', ` Darkness', ` Caps', ` blush', ` Highest'} & \textbf{0.0040} \\
\multirow{6}{*}{\includegraphics[width=0.115\linewidth]{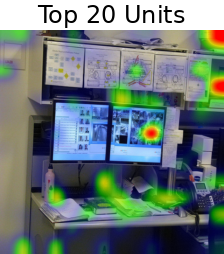}} & L9.u3561 & 93 & ` mix', ` CRC', ` critically', ` gulf', ` mechanically' & 0.0040 \\
 & \textbf{L7.u5533} & \textbf{93} & \textbf{`bags', `Items', ` comprehens', ` decor', `bag'} & \textbf{0.0039} \\
 & \textbf{L9.u8687} & \textbf{93} & \textbf{` eaten', ` foods', ` food', ` diet', ` eating'} & \textbf{0.0037} \\
 & \textbf{L12.u4109} & \textbf{93} & \textbf{` Lakes', ` Hof', ` Kass', ` Cotton', `Council'} & \textbf{0.0036} \\
 & \textbf{L8.u943} & \textbf{93} & \textbf{` Foods', `Food', `let', ` lunch', `commercial'} & \textbf{0.0036} \\
 & \textbf{L5.u16106} & \textbf{93} & \textbf{`ware', ` halls', ` salt', `WARE', ` mat'} & \textbf{0.0032} \\
 & \textbf{L8.u14504} & \textbf{143} & \textbf{` upstairs', ` homeowners', ` apartments', ` houses', ` apartment'} & \textbf{0.0032} \\
\multirow{6}{*}{\includegraphics[width=0.115\linewidth]{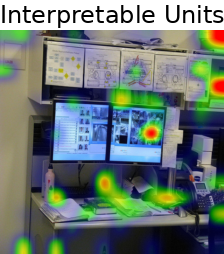}} & \textbf{L9.u11735} & \textbf{93} & \textbf{` hysterical', ` Gould', ` Louie', ` Gamble', ` Brown'} & \textbf{0.0031} \\
 & \textbf{L8.u14504} & \textbf{149} & \textbf{` upstairs', ` homeowners', ` apartments', ` houses', ` apartment'} & \textbf{0.0031} \\
 & \textbf{L5.u2771} & \textbf{93} & \textbf{` occupations', ` industries', ` operations', ` occupational', ` agriculture'} & \textbf{0.0029} \\
 & L9.u15864 & 55 & `lihood', `/**', `Advertisements', `.''.', `"""' & 0.0028 \\
 & \textbf{L9.u4691} & \textbf{149} & \textbf{` houses', ` buildings', ` dwellings', ` apartments', ` homes'} & \textbf{0.0028} \\
 & \textbf{L7.u10853} & \textbf{13} & \textbf{` boutique', ` firm', ` Associates', ` restaurant', ` Gifts'} & \textbf{0.0028} \\
\midrule
\multirow{6}{*}{\includegraphics[width=0.115\linewidth]{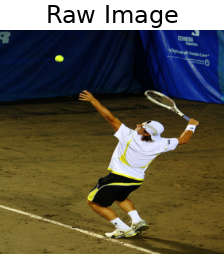}} & \textbf{L8.u15435} & \textbf{160} & \textbf{` tennis', ` tournaments', ` tournament', ` golf', ` racing'} & \textbf{0.0038} \\
 & L1.u15996 & 132 & `276', `PS', `ley', `room', ` Will' & 0.0038 \\
 & L5.u6439 & 160 & ` ge', ` fibers', ` hair', ` geometric', ` ori' & 0.0037 \\
 & L9.u15864 & 160 & `lihood', `/**', `Advertisements', `.''.', `"""' & 0.0034 \\
 & L12.u2955 & 160 & `Untitled', `Welcome', `========', `Newsletter', `====' & 0.0033 \\
 & L12.u2955 & 146 & `Untitled', `Welcome', `========', `Newsletter', `====' & 0.0032 \\
 & L7.u2688 & 160 & `rection', `itud', ` Ratio', `lat', ` ratio' & 0.0031 \\
\multirow{6}{*}{\includegraphics[width=0.115\linewidth]{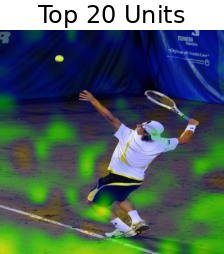}} & \textbf{L8.u4372} & \textbf{160} & \textbf{` footage', ` filmed', ` filming', ` videos', ` clips'} & \textbf{0.0029} \\
 & L10.u4819 & 146 & `çĶ°', `¬¼', `"""', ` Marketable', `å§' & 0.0029 \\
 & \textbf{L8.u15435} & \textbf{93} & \textbf{` tennis', ` tournaments', ` tournament', ` golf', ` racing'} & \textbf{0.0029} \\
 & \textbf{L8.u15435} & \textbf{146} & \textbf{` tennis', ` tournaments', ` tournament', ` golf', ` racing'} & \textbf{0.0029} \\
 & L10.u927 & 132 & `onds', `rog', `lys', `arrow', `ond' & 0.0027 \\
 & L9.u15864 & 146 & `lihood', `/**', `Advertisements', `.''.', `"""' & 0.0026 \\
 & L1.u8731 & 132 & ` âĢ¦', ` [âĢ¦]', `âĢ¦', ` ...', ` Will' & 0.0025 \\
\multirow{6}{*}{\includegraphics[width=0.115\linewidth]{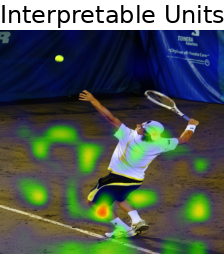}} & \textbf{L8.u16330} & \textbf{160} & \textbf{` bouncing', ` hitting', ` bounce', ` moving', ` bounced'} & \textbf{0.0025} \\
 & L9.u1908 & 146 & ` members', ` country', ` VIII', ` Spanish', ` 330' & 0.0024 \\
 & L10.u4819 & 160 & `çĶ°', `¬¼', `"""', ` Marketable', `å§' & 0.0024 \\
 & \textbf{L11.u14710} & \textbf{160} & \textbf{`Search', `Follow', `Early', `Compar', `Category'} & \textbf{0.0024} \\
 & L6.u132 & 160 & ` manually', ` replace', ` concurrently', `otropic', ` foregoing' & 0.0024 \\
 & \textbf{L7.u5002} & \textbf{160} & \textbf{` painting', ` paintings', ` sculpture', ` sculptures', ` painted'} & \textbf{0.0024} \\
\midrule
\end{tabularx}
\end{table*}

\begin{table*}[htbp]
\centering
\small
\label{table:example3}
\begin{tabularx}{\linewidth}{c l l p{9.6cm} r}
\toprule
\textbf{Images} & \textbf{Layer.unit} & \textbf{Patch} & \textbf{Decoding (top 5 tokens)} & \textbf{Attr. score} \\
\midrule
\multirow{6}{*}{\includegraphics[width=0.115\linewidth]{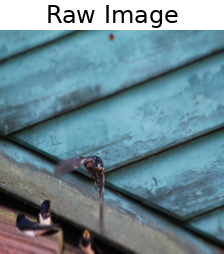}} & \textbf{L5.u13680} & \textbf{132} & \textbf{` driver', ` drivers', ` cars', `heading', `cars'} & \textbf{0.0091} \\
 & \textbf{L11.u9566} & \textbf{132} & \textbf{` traffic', ` network', ` networks', ` Traffic', `network'} & \textbf{0.0090} \\
 & \textbf{L12.u11606} & \textbf{132} & \textbf{` chassis', ` automotive', ` design', ` electronics', ` specs'} & \textbf{0.0078} \\
 & \textbf{L7.u6109} & \textbf{132} & \textbf{` automobile', ` automobiles', ` engine', ` Engine', ` cars'} & \textbf{0.0078} \\
 & \textbf{L6.u11916} & \textbf{132} & \textbf{` herd', `loads', ` racing', ` herds', ` horses'} & \textbf{0.0071} \\
 & \textbf{L8.u562} & \textbf{132} & \textbf{` vehicles', ` vehicle', ` cars', `veh', ` Vehicles'} & \textbf{0.0063} \\
 & \textbf{L7.u3273} & \textbf{132} & \textbf{`ride', ` riders', ` rides', ` ridden', ` rider'} & \textbf{0.0062} \\
\multirow{6}{*}{\includegraphics[width=0.115\linewidth]{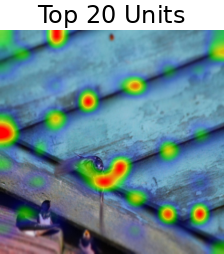}} & \textbf{L13.u5734} & \textbf{132} & \textbf{` Chevrolet', ` Motorsport', ` cars', ` automotive', ` vehicle'} & \textbf{0.0062} \\
 & \textbf{L8.u2952} & \textbf{132} & \textbf{` rigging', ` valves', ` nozzle', ` pipes', ` tubing'} & \textbf{0.0059} \\
 & \textbf{L13.u8962} & \textbf{132} & \textbf{` cruising', ` flying', ` flight', ` refuel', ` Flying'} & \textbf{0.0052} \\
 & L9.u3561 & 116 & ` mix', ` CRC', ` critically', ` gulf', ` mechanically' & 0.0051 \\
 & \textbf{L13.u107} & \textbf{132} & \textbf{` trucks', ` truck', ` trailer', ` parked', ` driver'} & \textbf{0.0050} \\
 & \textbf{L14.u10852} & \textbf{132} & \textbf{`Veh', ` driver', ` automotive', ` automakers', `Driver'} & \textbf{0.0049} \\
 & L6.u1989 & 132 & `text', `light', `TL', `X', `background' & 0.0049 \\
\multirow{6}{*}{\includegraphics[width=0.115\linewidth]{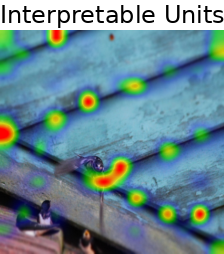}} & L2.u14243 & 132 & `ousel', ` Warriors', `riages', `illion', `Ord' & 0.0048 \\
 & \textbf{L5.u6589} & \textbf{132} & \textbf{` vehicles', ` motorcycles', ` aircraft', ` tyres', ` cars'} & \textbf{0.0046} \\
 & \textbf{L7.u4574} & \textbf{132} & \textbf{` plants', ` plant', ` roof', ` compost', ` wastewater'} & \textbf{0.0045} \\
 & \textbf{L7.u6543} & \textbf{132} & \textbf{` distance', ` downhill', ` biking', ` riders', ` journeys'} & \textbf{0.0045} \\
 & \textbf{L16.u9154} & \textbf{132} & \textbf{` driver', ` drivers', ` vehicle', ` vehicles', `driver'} & \textbf{0.0045} \\
 & \textbf{L12.u7344} & \textbf{132} & \textbf{` commemor', ` streets', ` celebrations', ` Streets', ` highways'} & \textbf{0.0044} \\
\midrule
\multirow{6}{*}{\includegraphics[width=0.115\linewidth]{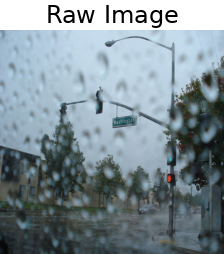}} & \textbf{L12.u9058} & \textbf{174} & \textbf{` swimming', ` Swim', ` swim', ` fishes', ` water'} & \textbf{0.0062} \\
 & \textbf{L17.u10507} & \textbf{174} & \textbf{` rivers', ` river', ` lake', ` lakes', ` River'} & \textbf{0.0049} \\
 & \textbf{L7.u3138} & \textbf{174} & \textbf{` basin', ` ocean', ` islands', ` valleys', ` mountains'} & \textbf{0.0046} \\
 & \textbf{L5.u6930} & \textbf{149} & \textbf{` rivers', ` river', ` River', ` waters', ` waterways'} & \textbf{0.0042} \\
 & \textbf{L7.u14218} & \textbf{174} & \textbf{` docks', ` Coast', ` swimming', ` swim', `melon'} & \textbf{0.0040} \\
 & \textbf{L9.u4379} & \textbf{149} & \textbf{` river', ` stream', ` River', ` Valley', ` flow'} & \textbf{0.0038} \\
 & \textbf{L6.u5868} & \textbf{149} & \textbf{`water', ` water', ` waters', ` river', ` River'} & \textbf{0.0036} \\
\multirow{6}{*}{\includegraphics[width=0.115\linewidth]{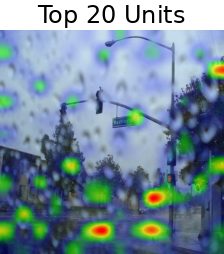}} & \textbf{L9.u4379} & \textbf{174} & \textbf{` river', ` stream', ` River', ` Valley', ` flow'} & \textbf{0.0036} \\
 & \textbf{L5.u6930} & \textbf{174} & \textbf{` rivers', ` river', ` River', ` waters', ` waterways'} & \textbf{0.0032} \\
 & \textbf{L7.u3138} & \textbf{149} & \textbf{` basin', ` ocean', ` islands', ` valleys', ` mountains'} & \textbf{0.0029} \\
 & \textbf{L6.u5868} & \textbf{174} & \textbf{`water', ` water', ` waters', ` river', ` River'} & \textbf{0.0028} \\
 & L7.u416 & 136 & ` praise', ` glimpse', ` glimps', ` palate', ` flavours' & 0.0027 \\
 & \textbf{L10.u15235} & \textbf{149} & \textbf{` water', ` waters', `water', ` lake', ` lakes'} & \textbf{0.0026} \\
 & \textbf{L4.u2665} & \textbf{136} & \textbf{` levels', ` absorbed', ` density', ` absorption', ` equilibrium'} & \textbf{0.0026} \\
\multirow{6}{*}{\includegraphics[width=0.115\linewidth]{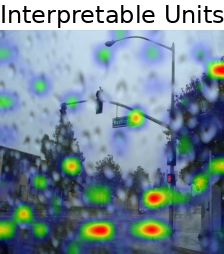}} & \textbf{L10.u14355} & \textbf{149} & \textbf{` roads', ` paths', ` flows', ` routes', ` streams'} & \textbf{0.0026} \\
 & \textbf{L17.u10507} & \textbf{149} & \textbf{` rivers', ` river', ` lake', ` lakes', ` River'} & \textbf{0.0024} \\
 & \textbf{L7.u7669} & \textbf{174} & \textbf{` weather', ` season', ` forecast', ` rains', ` winters'} & \textbf{0.0024} \\
 & \textbf{L8.u9322} & \textbf{136} & \textbf{` combustion', ` turbulence', ` recoil', ` vibration', ` hydrogen'} & \textbf{0.0024} \\
 & L9.u15864 & 182 & `lihood', `/**', `Advertisements', `.''.', `"""' & 0.0022 \\
 & \textbf{L7.u3138} & \textbf{78} & \textbf{` basin', ` ocean', ` islands', ` valleys', ` mountains'} & \textbf{0.0021} \\
\midrule
\end{tabularx}
\vspace{2mm}
\caption{Results of attribution analysis for randomly sampled images from the COCO validation set. Includes decoded tokens for the top 20 units by attribution score. The first column shows the COCO image and superimposed heatmaps of the mean activations from the top 20 units and the top interpretable units (shown in \textbf{bold}). Units can repeat if they attain a high attribution score on multiple image patches.}
\label{tab:randomimages}
\vspace{20mm}
\end{table*}


\begin{table*}[t!]
\begin{centering}
\small
\begin{tabular}{l@{\hspace{25pt}} l@{\hspace{25pt}} l l l}
\toprule
\textbf{Image} & \textbf{COCO Human Captions} & \textbf{GPT Caption}  & &\\ \midrule
\multirow{5}{*}{\includegraphics[width=0.14\textwidth]{figures/snowboarder.jpeg}} & 
A man riding a snowboard down the side of a snow covered slope. & A person jumping on the ice. & &  \\
&  A man snowboarding down the side of a snowy mountain. & & &  \\
&  Person snowboarding down a steep snow covered slope. & & &  \\
&  A person snowboards on top of a snowy path.& & &  \\
& The person holds both hands in the air while snowboarding.& & & \\
\end{tabular}
\end{centering}
\end{table*}

\begin{table*}[t!]
\begin{centering}
\small
\label{tab:supp_embeddings}
\begin{tabular}{l l l l l}
\toprule
\textbf{Patch} & \textbf{Image soft prompt (nearest neighbor tokens)} & \textbf{BSc.} & \textbf{Random soft prompt (nearest neighbor tokens)} &\textbf{BSc.} \\ 
\midrule
144 & [`nav', `GY', `+++', `done', `Sets'] & .29 &  [`Movement', `Ord', `CLUD', `levy', `LI'] & .31 \\
80 & [`heels', `merits', `flames', `platform', `fledged'] & .36 & [`adic', `Stub', `imb', `VER', `stroke'] & .34  \\
169 & [`ear', `Nelson', `Garden', `Phill', `Gun'] & .32 &  [`Thank', `zilla', `Develop', `Invest', `Fair'] & .31 \\
81 & [`vanilla', `Poc', `Heritage', `Tarant', `bridge'] & .33 & [`Greek', `eph', `jobs', `phylogen', `TM'] & .30 \\
89 & [`oily', `stant', `cement', `Caribbean', `Nad'] &  .37 & [`Forestry', `Mage', `Hatch', `Buddh', `Beaut'] & .34 \\
124 & [`ension', `ideas', `GY', `uler', `Nelson'] & .32 & [`itone', `gest', `Af', `iple', `Dial'] & .30 \\
5 & [`proves', `Feed', `meaning', `zzle', `stripe'] & .31 & [`multitude', `psychologically', `Taliban', `Elf', `Pakistan'] & .36 \\
175 & [`util', `elson', `asser', `seek', `////////////////////'] & .26 &  [`ags', `Git', `mm', `Morning', `Cit'] & .33 \\
55 & [`Judicial', `wasting', `oen', `oplan', `trade'] & .34 & [`odd', `alo', `rophic', `perv', `pei'] & .34 \\
61 & [`+++', `DEP', `enum', `vernight', `posted'] & .33 & [`Newspaper', `iii', `INK', `Graph', `UT'] & .35 \\
103 & [`Doc', `Barth', `details', `DEF', `buckets'] & .34 &[`pleas', `Eclipse', `plots', `cb', `Menu'] & .36 \\
99 & [`+++', `Condition', `Daytona', `oir', `research'] & .35 & [`Salary', `card', `mobile', `Cour', `Hawth'] & .35 \\
155 & [`Named', `910', `collar', `Lars', `Cats'] & .33 &[`Champ', `falsely', `atism', `styles', `Champ']& .30 \\
145 & [`cer', `args', `olis', `te', `atin'] & .30 & [`Chuck', `goose', `anthem', `wise', `fare']& .33 \\
189 & [`MOD', `Pres', `News', `Early', `Herz'] & .33 & [`Organ', `CES', `POL', `201', `Stan'] & .31 \\
49 & [`Pir', `Pir', `uum', `akable', `Prairie'] & .30 & [`flame', `roc', `module', `swaps', `Faction'] & .33 \\
20 & [`ear', `feed', `attire', `demise', `peg'] & .33 & [`Chart', `iw', `Kirst', `PATH', `rhy'] & .36 \\ 
110 & [`+++', `Bee', `limits', `Fore', `seeking'] & .31 & [`imped', `iola', `Prince', `inel', `law'] & .33 \\
6 & [`SIGN', `Kob', `Ship', `Near', `buzz'] & .36 & [`Tower', `767', `Kok', `Tele', `Arbit'] & .33 \\
46 & [`childhood', `death', `ma', `vision', `Dire'] & .36 & [`Fram', `exper', `Pain', `ader', `unprotected'] & .33 \\
113 & [`Decl', `Hide', `Global', `orig', `meas'] & .32 & [`usercontent', `OTUS', `Georgia', `ech', `GRE']& .32\\
32 & [`ideas', `GY', `+++', `Bake', `Seed'] & .32 & [`GGGGGGGG', `dictators', `david', `ugh', `BY'] & .31 \\
98 & [`Near', `Near', `LIN', `Bee', `threat'] & .30 & [`Lavrov', `Debor', `Hegel', `Advertisement', `iak'] & .34 \\
185 & [`ceans', `Stage', `Dot', `Price', `Grid'] & .33 & [`wholesale', `Cellular', `Magn', `Ingredients', `Magn'] & .32 \\ 
166 & [`bys', `767', `+++', `bottles', `gif'] & .32 & [`Bras', `discipl', `gp', `AR', `Toys'] & .33 \\
52 & [`Kob', `Site', `reed', `Wiley', `âĻ'] & .29 & [`THER', `FAQ', `ibility', `ilities', `twitter'] & .34 \\
90 & [`cytok', `attack', `Plug', `strategies', `uddle'] & .32 & [`Boots', `Truman', `CFR', `ãĤ£', `Shin'] & .33 \\
13 & [`nard', `Planetary', `lawful', `Court', `eman'] & .33 & [`Nebraska', `tails', `ÅŁ', `DEC', `Despair']& .33 \\
47 & [`pport', `overnight', `Doc', `ierra', `Unknown'] & .34 & [`boiling', `A', `Ada', `itude', `flawed'] & .31 \\
19 & [`mocking', `chicks', `GY', `ear', `done'] & .35 & [`illet', `severely', `nton', `arrest', `Volunteers'] & .33 \\
112 & [`avenue', `gio', `Parking', `riages', `Herald'] & .35 & [`griev', `Swanson', `Guilty', `Sent', `Pac'] & .32 \\ 
133 & [`ãĤĬ', `itto', `iation', `asley', `Included'] & .32 & [`Purs', `reproductive', `sniper', `instruct', `Population'] & .33 \\
102 & [`drawn', `Super', `gency', `Type', `blames'] & .33 & [`metric', `Young', `princip', `scal', `Young'] & .31 \\
79 & [`Vand', `inement', `straw', `ridiculous', `Chick'] & .34 & [`Rez', `song', `LEGO', `Login', `pot'] & .37 \\
105 & [`link', `ede', `Dunk', `Pegasus', `Mao'] & .32 & [`visas', `Mental', `verbal', `WOM', `nda'] & .30 \\
\midrule
& \textbf{Average} & .33 & & .33\\
\bottomrule
\end{tabular}
\end{centering}
\vspace{2mm}
\caption{Image soft prompts are indistinguishable from random soft prompts via BERTScore. Each image is encoded as a sequence of 196 soft prompts, corresponding to image patches, that serve as input to GPT-J. Here we randomly sample 35 patches for a single COCO image and map them onto nearest-neighbor tokens in transformer embedding space. BERTScore is measured relative to COCO human annotations of the same image (we report the mean score over the $5$ human captions). For comparison we sample random vectors in the transformer embedding space and compute BERTScores using the same procedure.}
\end{table*}

\begin{figure*}[t!]
\begin{centering}
\includegraphics[width=\linewidth]{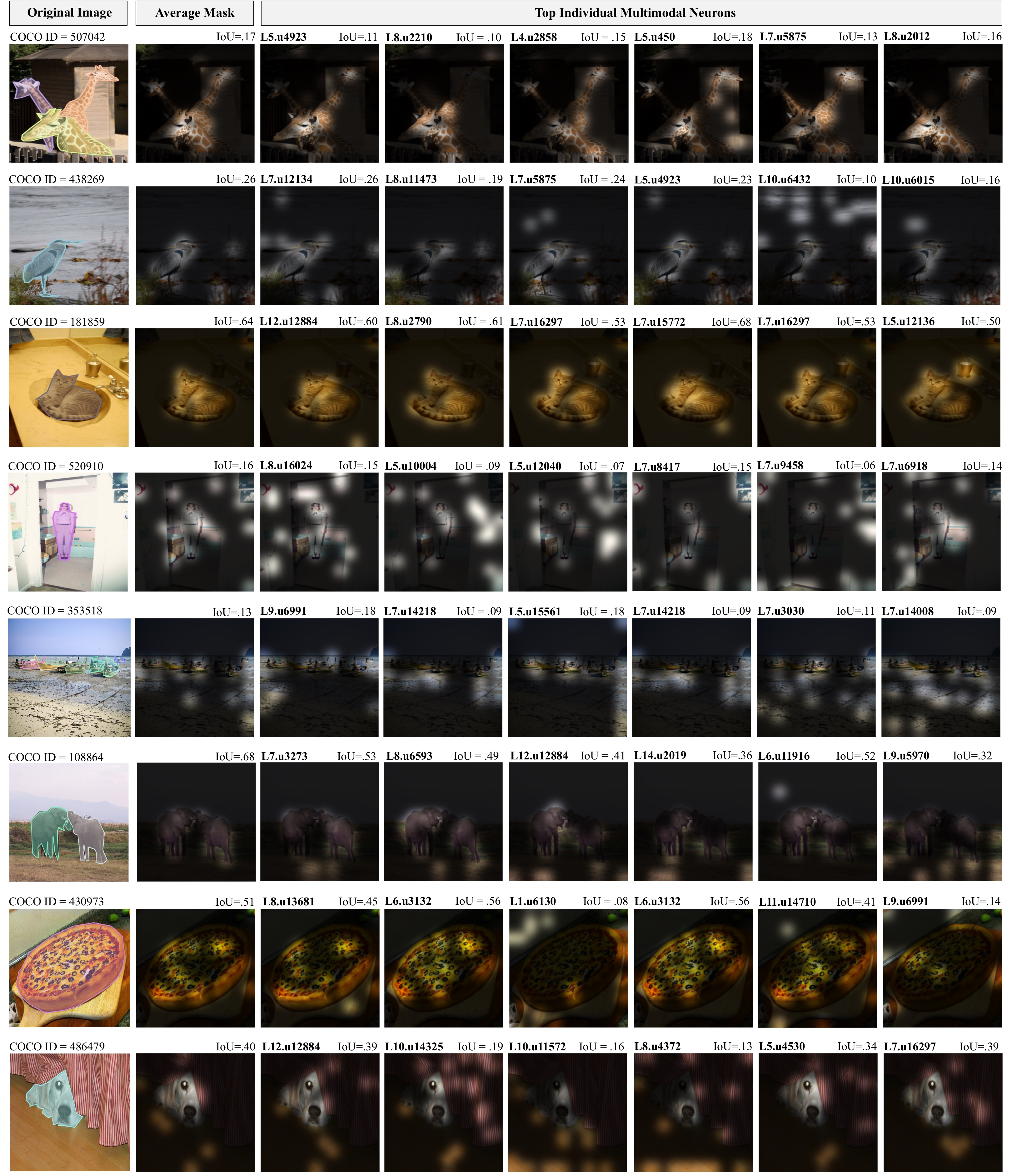}
\caption{Multimodal neurons are selective for objects in images. For $8$ example images sampled from the COCO categories described in Section 3.2 of the main paper, we show activation masks of individual multimodal neurons over the image, as well as mean activation masks over all top multimodal neurons. We use IoU to compare these activation masks to COCO object annotations. IoU is calculated by upsampling each activation mask to the size of the original image $(224)$ using bilinear interpolation, and thresholding activations in the 0.95 percentile to produce a binary segmentation mask.}
\end{centering}
\label{fig:IOUsupp}
\end{figure*}

\clearpage
\clearpage
\newpage

\begin{figure*}[hbt!]
\includegraphics[width=0.9\linewidth]{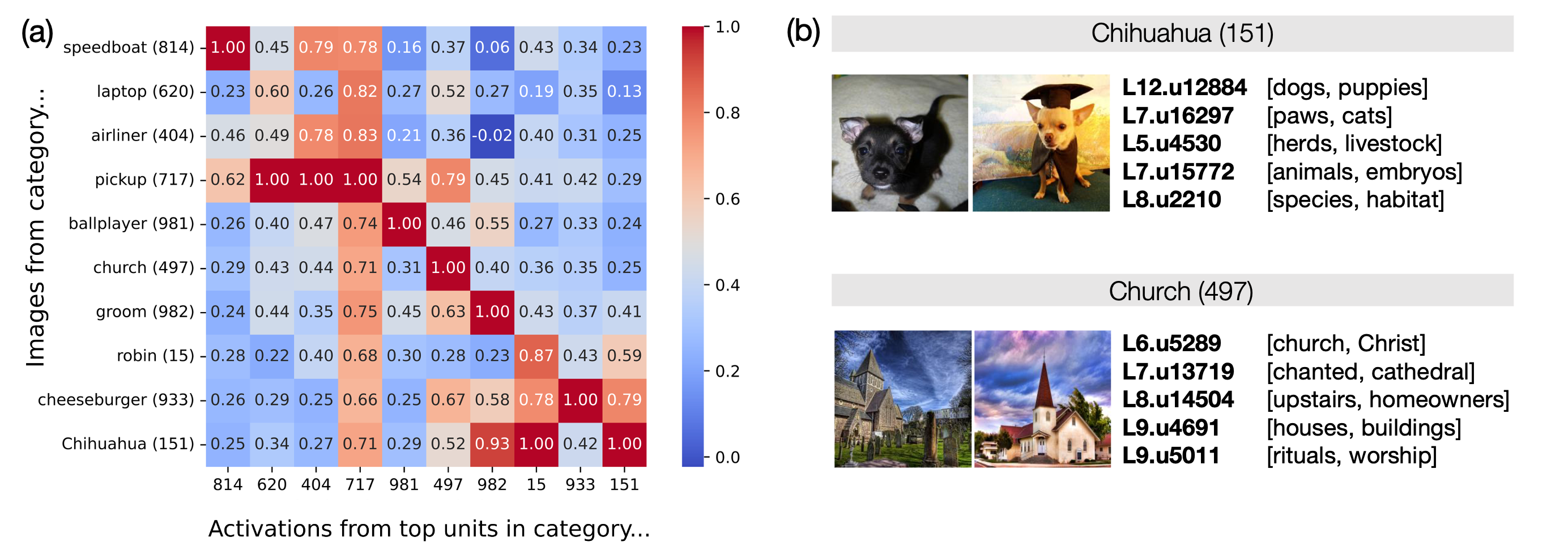}
   \caption{Multimodal neurons are selective for image categories. (a) For 10 ImageNet classes we construct the set of interpretable multimodal neurons with the highest attribution scores on training images in that class, and calculate their activations on validation images. For each class, we report the average activation value of top-scoring multimodal units relative to the maximum value of their average activations on any class. Multimodal neurons are maximally active on classes where their attribution scores are highest. (b) Sample images and top-scoring units from two classes.}
   \vspace{-4mm}
\label{fig:confusionmatrix}
\end{figure*}

\clearpage 
\newpage

\section{Ablating Multimodal Neurons}
\noindent
In Section 3.3 of the main paper, we show that ablating multimodal neurons causally effects the probability of outputting the original token. To investigate the effect of removing multimodal neurons on model output, we ablate the top $k$ units by attribution score for an image, where $k \in \{0, 50, 100, 200, 400, 800, 1600, 3200, 6400\},$ and compute the BERTScore between the model's original caption and the newly-generated zero-temperature caption. Whether we remove the top $k$ units by attribution score, or only those that are interpretable, we observe a strong decrease in caption similarity. Table S.3 shows examples of the effect of ablating top neurons on randomly sampled COCO validation images, compared to the effect of ablating random neurons. Figure S.3 shows the average BERTScore after ablating $k$ units across all COCO validation images. 

\begin{figure}[hbt!]
\includegraphics[width=\linewidth]{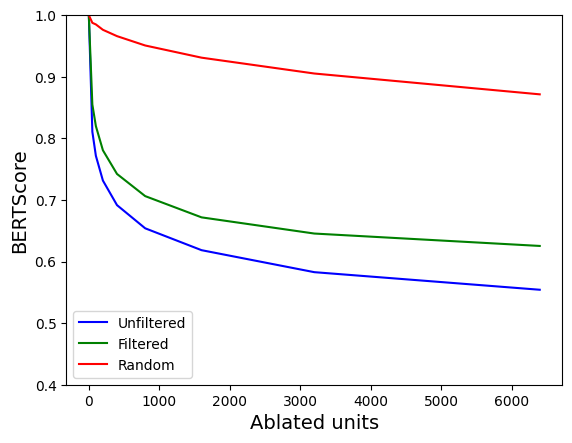}
\caption{BERTScores of generated captions decrease when multimodal neurons are ablated, compared to the ablation of random neurons from the same layers. }
\label{fig:ablationbert}
\end{figure}

\newpage

\section{Distribution of Multimodal Neurons}
\noindent
We perform a simple analysis of the distribution of multimodal neurons by layer. Specifically, we extract the top 100 scoring neurons for all COCO validation images. Most of these neurons are found between layers $5$ and $10$ of GPT-J ($L=28$), suggesting translation of semantic content between modalities occurs in earlier transformer layers.

\begin{figure}[hbt!]
\begin{center}
\includegraphics[width=\linewidth]{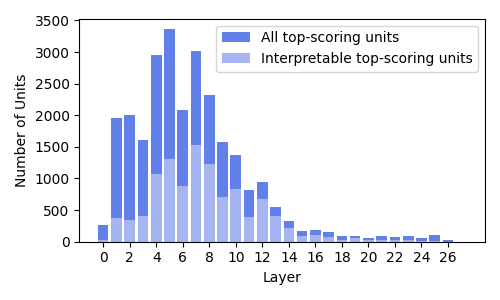}
   \caption{Unique multimodal neurons per layer chosen using the top 100 attribution scores for each COCO validation image. Interpretable units are those for which at least 7 of the top 10 logits are words in the English dictionary containing $\geq 3$ letters.}
   \label{fig:layerunits}
   \vspace{-8mm}
\end{center}
\end{figure}

\clearpage
\newpage

\begin{figure*}[t!]
\begin{center}
\includegraphics[width=.99\linewidth]{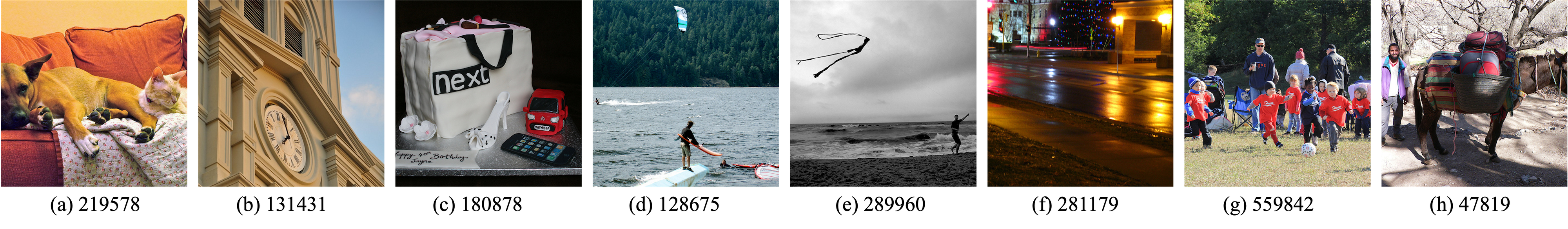}
\vspace{-6mm}
\end{center}
\end{figure*}

\begin{table*}[b!]
\centering
\small
\label{table:ablations}
\begin{tabular}{l l l l l l l l}
\toprule
& & \multicolumn{5}{c}{\normalsize{\textbf{Captions after ablation}}} \\
\cmidrule{3-8}
\small{\textbf{Img. ID}} & \small{\textbf{$\#$ Abl.}} & \multicolumn{1}{l}{\textbf{All multimodal}} & \multicolumn{1}{c}{ BSc.}& \multicolumn{1}{l}{\textbf{Interpretable multimodal}} & \multicolumn{1}{c}{BSc.} & \multicolumn{1}{l}{\textbf{Random neurons}} & \multicolumn{1}{c}{BSc.}\\
\midrule

\small{219578} & 0 & a dog with a cat & 1.0 & a dog with a cat & 1.0  & a dog with a cat & 1.0  \\
 & 50 & a dog and a cat & .83  & a dog and a cat & .83  & a dog with a cat & 1.0  \\
 & 100 & a lion and a zebra & .71  & a dog and cat & .80  & a dog with a cat & 1.0  \\
 & 200 & a dog and a cat & .83  & a dog and a cat & .83  & a dog with a cat & 1.0  \\
& 400 & a lion and a lioness & .64  & a dog and a cat & .83  & a dog with a cat  & 1.0 \\
& 800 & a tiger and a tiger & .63  & a lion and a zebra & .71  & a dog with a cat & 1.0  \\
& 1600 & a tiger and a tiger & .63  & a lion and a zebra & .71  & a dog with a cat & 1.0  \\
& 3200 & a tiger  & .67 & a tiger and a tiger & .63 & a dog with a cat & 1.0  \\
& 6400 & a tiger & .67  & a tiger in the jungle & .60  & a dog with a cat & 1.0  \\
\midrule
\small{131431} & 0 & the facade of the cathedral  & 1.0 & the facade of the cathedral & 1.0  & the facade of the cathedral & 1.0  \\

 & 50 & the facade of the church & .93  & the facade of the cathedral & 1.0 & the facade of the cathedral & 1.0  \\
 
 & 100 & the facade of the church &  .93 & the facade of the cathedral & 1.0  & the facade of the cathedral & 1.0  \\
 
 & 200 & the facade & .75  & the facade & .75  & the facade of the cathedral & 1.0  \\
 
& 400 & the exterior of the church & .80  & the facade & .75  & the facade of the cathedral  & 1.0 \\

& 800 & the exterior of the church & .80  & the dome & .65  & the facade of the cathedral & 1.0  \\

& 1600 & the dome & .65  & the dome & .65  & the facade of the cathedral & 1.0  \\

& 3200 & the dome & .65 & the dome & .65  & the facade of the cathedral & 1.0  \\

& 6400 & the exterior & .61  & the dome & .65  & the facade & .75  \\

\midrule
\small{180878} & 0 & a cake with a message &  & a cake with a message  & & a cake with a message &   \\
& & written on it. & 1.0 & written on it. & 1.0 & written on it. & \small{1.0}\\

 & 50 & a cake with a message  &  & a cake with a message  & & a cake with a message &   \\
& & written on it. & 1.0 & written on it. & 1.0 & written on it. & \small{1.0}\\

 & 100 & a cake with a message  &  & a cake for a friend's birthday. & .59 & a cake with a message &   \\
& & written on it. & 1.0 & &  & written on it. & \small{1.0}\\

 & 200 & a cake with a message  &  & a cake for a friend's birthday.  & .59 & a cake with a message &   \\
& & written on it. & 1.0 &  &  & written on it. & \small{1.0}\\

 & 400 & a cake with a message  &  & a cake for a friend's birthday.  & .59 & a cake with a message &   \\
& & written on it. & 1.0 &  &  & written on it. & \small{1.0}\\

 & 800 & a cake  & .59  & a cake for a birthday party  & .56 & a cake with a message &   \\
& &  &  &  &  & written on it. & \small{1.0}\\

 & 1600 & a cake  & .59 & a poster for the film.  &  .49 & a cake with a message &   \\
& &  &  & & & written on it. & \small{1.0}\\

 & 3200 & a man who is a fan of  &  & a typewriter  & .44 & a cake with a message &   \\
& & football & .42 &  & & written on it. & \small{1.0}\\

 & 6400 & the day  & .34 & a typewriter  & .44 & a cake with a message &   \\
& &  &  &  & & written on it. & \small{1.0}\\

\midrule

\small{128675} & 0 & \small{a man surfing on a wave} & \small{1.0} & a man surfing on a wave & 1.0  & a man surfing on a wave & \small{1.0}  \\

 & 50 & \small{a man in a kayak on a lake} & .74  & a man surfing on a wave & 1.0  & a man surfing on a wave & \small{1.0}  \\
 
 & 100 & a man in a kayak on a lake & .74  & a man surfing on a wave & 1.0  & a man surfing on a wave & \small{1.0}  \\
 
 & 200 & a man in a kayak on a lake & .74  & a man surfing a wave & .94  & a man surfing on a wave & \small{1.0}  \\
 
& 400 & a man in a kayak on a lake & .74  & a man surfing a wave & .94  & a man surfing on a wave  & \small{1.0} \\

& 800 & a man in a kayak & .64  & a surfer riding a wave & .84  & a man surfing on a wave & \small{1.0} \\

& 1600 & a girl in a red dress &  & a surfer riding a wave & .84  & a man surfing on a wave & \small{1.0}  \\
&  & walking on the beach & .66  &  &  &   &   \\

& 3200 & a girl in a red dress  & .53 & a girl in a red dress & .53  & a man surfing on a wave & \small{1.0}  \\

& 6400 & a girl in the water & .62  & a girl in a dress & .59 & a man surfing on a wave & \small{1.0}  \\

\bottomrule
\end{tabular}
\end{table*}

\begin{table*}[htbp]
\centering
\small
\label{table:ablations2}
\begin{tabular}{l l l l l l l l}
\toprule
\small{\textbf{Img. ID}} & \small{\textbf{$\#$ Abl.}} & \multicolumn{1}{l}{\textbf{All multimodal}} & \multicolumn{1}{c}{ BSc.}& \multicolumn{1}{l}{\textbf{Interpretable multimodal}} & \multicolumn{1}{c}{BSc.} & \multicolumn{1}{l}{\textbf{Random neurons}} & \multicolumn{1}{c}{BSc.}\\
\midrule

\small{289960} & 0 & a man standing on a rock &  & a man standing on a rock  & & a man standing on a rock &   \\
& & in the sea & 1.0 & in the sea & 1.0 & in the sea & \small{1.0}\\

 & 50 & a man standing on a rock  &  & a man standing on a rock  & & a man standing on a rock &   \\
& & in the sea & 1.0 & in the sea & 1.0 & in the sea & \small{1.0}\\

 & 100 & a man standing on a rock  &  & a man standing on a rock &  & a man standing on a rock &   \\
& & in the sea & 1.0 &in the sea. & .94 & in the sea & \small{1.0}\\

 & 200 & a kite soaring above the waves  & .62 & a man standing on a rock  &  & a man standing on a rock &   \\
& &  &  & in the sea & 1.0 & in the sea & \small{1.0}\\

 & 400 & a kite soaring above the waves  & .62 & a kite surfer on the beach.  & .62 & a man standing on a rock &   \\
& &  & &  &  & in the sea & \small{1.0}\\

 & 800 & a kite soaring above the waves  & .62  & a bird on a wire  & .63 & a man standing on a rock &   \\
& &  &  &  &  & in the sea & \small{1.0}\\

 & 1600 & a kite soaring above the clouds  & .65 & a kite surfer on the beach  &  .63 & a man standing on a rock &   \\
& &  &  & & & in the sea & \small{1.0}\\

 & 3200 & a kite soaring above the sea  & .69 & a bird on a wire  & .63 & a man standing on a rock &   \\
& & &  &  & & in the sea & \small{1.0}\\

 & 6400 & a helicopter flying over the sea  & .69 & a bird on a wire  & .63 & a man standing on a rock &   \\
& &  &  &  & & in the sea & \small{1.0}\\

\midrule
\small{131431} & 0 & the bridge at night  & 1.0 & the bridge at night & 1.0  & the bridge at night & 1.0  \\

 & 50 & the bridge & .70  & the street at night & .82 & the bridge at night & 1.0  \\
 
 & 100 & the bridge &  .70 & the street at night & .82  & the bridge at night & 1.0  \\
 
 & 200 & the bridge & .70  & the street at night & .82  & the bridge at night & 1.0  \\
 
& 400 & the bridge & .70  & the street & .55  & the bridge at night  & 1.0 \\

& 800 & the bridge & .70  & the street& .55  & the bridge at night & 1.0  \\

& 1600 & the bridge & .70  & the street & .55  & the bridge at night & 1.0  \\

& 3200 & the night & .61 & the street & .55  & the bridge at night & 1.0  \\

& 6400 & the night & .61 & the street & .55  & the bridge at night & 1.0  \\

\midrule
\small{559842} & 0 & the team during the match. & 1.0 & the team during the match. & 1.0  & the team during the match. & 1.0  \\
 & 50 & the team. & .70  & the team. & .70  & the team during the match. & 1.0  \\
 & 100 & the team. & .70  & the team. & .70  & the team during the match. & 1.0  \\
 & 200 & the team. & .70  & the team. & .70 & the team during the match. & 1.0  \\
& 400 & the group of people & .52  & the team. & .70  & the team during the match.  & 1.0 \\
& 800 & the group & .54  & the team. & .70 & the team during the match. & 1.0  \\
& 1600 & the group & .54  & the team. & .70 & the team during the match. & 1.0  \\
& 3200 & the group  & .54 & the team. & .70 & the team during the match & 1.0  \\

& 6400 & the kids & .46  & the team. & .70  & the team during the match. & 1.0  \\

\midrule

\small{47819} & 0 & a man and his horse. & \small{1.0} & a man and his horse. & 1.0  & a man and his horse. & \small{1.0}  \\

 & 50 & a man and his horse. & 1.0  & a man and his horse. & 1.0  & a man and his horse. & \small{1.0}  \\
 
 & 100 & the soldiers on the road & .47 & a man and his horse. & 1.0  & a man and his horse. & \small{1.0}  \\
 
 & 200 & the soldiers on the road & .47  & the soldiers on the road & .47 & a man and his horse. & \small{1.0}  \\
 
& 400 & the soldiers & .46 & the soldiers & .46  & a man and his horse.  & \small{1.0} \\

& 800 & the soldiers & .46  & the soldiers & .46  & a man and his horse. & \small{1.0} \\

& 1600 & the soldiers & .46 & the soldiers & .46  & a man and his horse. & \small{1.0}  \\

& 3200 & the soldiers  & .46 & the soldiers & .46  & a man and his horse. & \small{1.0}  \\

& 6400 & the soldiers & .46  & the soldiers & .46 & a man and his horse. & \small{1.0}  \\

\bottomrule
\end{tabular}
\vspace{2mm}
\caption{Captions and BERTScores (relative to original GPT caption) after incremental ablation of multimodal MLP neurons. All multimodal neurons are detected, decoded, and filtered to produce a list of ``interpretable'' multimodal neurons using the procedure described in Section 2 of the main paper. Random neurons are sampled from the same layers as multimodal neurons for ablation. Images are randomly sampled from the COCO validation set. Captions are generated with temperature $=0$.}
\end{table*}

\end{document}